\documentclass[10pt]{article}
\usepackage[preprint]{tmlr}

\usepackage[utf8]{inputenc} 
\usepackage[T1]{fontenc}    
\usepackage{hyperref}       
\usepackage{url}            
\usepackage{booktabs}       
\usepackage{amsfonts}       
\usepackage{nicefrac}       
\usepackage{microtype}      
\usepackage{xcolor}         
\usepackage{amsmath}
\usepackage{xcolor}
\usepackage{amsmath}
\usepackage{graphicx}
\usepackage{wrapfig}
\usepackage{bm}
\usepackage{pdflscape}
\usepackage{todonotes}
\usepackage{subcaption}
\usepackage[title]{appendix}
\usepackage{hyperref}
\usepackage{url}

\title{Uncertainty quantification and out-of-distribution detection using surjective normalizing flows}

\author{%
  Simon Dirmeier\thanks{Correspondence to: \texttt{simon.dirmeier@sdsc.ethz.ch}}\\
  \addr  Swiss Data Science Center\\
  ETH Zurich, Switzerland\\
  \AND
  Ye Hong\\
  \addr ETH Zurich, Switzerland\\
  \AND
  Yanan Xin\\
  \addr ETH Zurich, Switzerland\\
  \AND
  Fernando Perez-Cruz\\
  \addr Swiss Data Science Center\\
  ETH Zurich, Switzerland\\
}

\def\month{MM}  
\def\year{YYYY} 
\def\openreview{\url{https://openreview.net/forum?id=XXXX}} 

\begin{document}

\maketitle

\begin{abstract}
Reliable quantification of epistemic and aleatoric uncertainty is of crucial importance in applications where models are trained in one environment but applied to multiple different environments, often seen in real-world applications for example, in climate science or mobility analysis. We propose a simple approach using surjective normalizing flows to identify out-of-distribution data sets in deep neural network models that can be computed in a single forward pass. The method builds on recent developments in deep uncertainty quantification and generative modeling with normalizing flows. We apply our method to a synthetic data set that has been simulated using a mechanistic model from the mobility literature and several data sets simulated from interventional distributions induced by soft and atomic interventions on that model, and demonstrate that our method can reliably discern out-of-distribution data from in-distribution data. We compare the surjective flow model to a Dirichlet process mixture model and a bijective flow and find that the surjections are a crucial component to reliably distinguish in-distribution from out-of-distribution data.
\end{abstract}

\section{Introduction}
Uncertainty quantification~\citep{lakshminarayanan2017simple,liu2020Simple,tagasovska2019single,amersfoort2020uncertainty,kong2020sde,mukhoti2021deterministic} and distributional robustness~\citep{peters2016causal,meinshausen2018causality,arjovsky2019invariant,heinze2021conditional,rothenhausler2021anchor} are increasingly being recognized as mandatory features of machine and deep learning methods to find successful applications in industrial or academic settings. For instance, in fields such as climate science or mobility analysis, it is often necessary to be able to train models in one environment, such as the traffic system of a city or the mobility behavior of an individual, and then deploy them in another environment~\citep{xin2022vision}. Similarly, when a prediction has been made, quantification of predictive uncertainty or confidence levels is often essential for reliable downstream decision-making. 

In scenarios where decisions need to be taken live, i.e., in location prediction for individual mobility or stock market forecasting in finance, uncertainty estimates are ideally calculated during the evaluation of a data point by a model, e.g., during the same forward pass of a neural network, which forbids computationally intensive approaches (such as \cite{lakshminarayanan2017simple} or \cite{tagasovska2019single}). Furthermore, due to the scarcity of data sets in privacy-sensitive disciplines like mobility research which often originate from confidential sources such as user surveys or GPS logs tracking user positions, the necessity arises for a suitable methodology that enables the computation of uncertainty estimates using the same predictive model.

In this work, we build on the work of \cite{mukhoti2021deterministic} and propose an approach to quantify aleatoric and epistemic uncertainty for deep learning models to detect out-of-distribution data. The method first computes epistemic uncertainty estimates using surjective normalizing flows (SNFs; \cite{nielsen2020survae,klein2021funnels,dirmeier2023simulation}) and detects if a data point is out-of-distribution if there is a significant difference to the density estimates of the training data assuming that low density of a data point is an indicator of being out-of-distribution. If the data is in-distribution, aleatoric density estimates can then be computed using the softmax entropy when outputs are discrete or the variance of the prediction when outputs are continuous.

The manuscript is organized as follows. Section~\ref{sec:methods} introduces the required background and a method that uses SNFs for epistemic uncertainty estimation, specifically, how neural network models can be extended to allow for epistemic uncertainty estimation using SNFs. Section~\ref{sec:results} showcases a real-world application from mobility research and evaluates the method in comparison to other density estimators. Section~\ref{sec:conclusion} summarizes our findings and discusses future avenues for research.

\section{Methods}
\label{sec:methods}
\subsection{Uncertainty quantification}

When quantifying uncertainty in scientific applications, it is often required to distinguish \textit{epistemic} from \textit{aleatoric} uncertainty \citep{hullermeier2021aleatoric}. We denote epistemic uncertainty as uncertainty that stems from insufficient amounts of training data which as a consequence manifests in uncertainty in parameter estimates. Hence, epistemic uncertainty is inversely proportional to the density of data samples and can be used for, e.g., detection of out-of-distribution samples. We denote aleatoric uncertainty as uncertainty of an estimator that is due to, for instance, measurement noise which cannot be reduced by increasing the sample size. Estimates of aleatoric uncertainty are high for ambiguous samples, e.g., if the same feature has been observed for different responses.

\subsection{Feature-space regularisation}
\label{sec:regularization}

\cite{mukhoti2021deterministic} demonstrate how neural network models can be used for epistemic and aleatoric uncertainty estimation. Specifically, they propose to regularize the feature-space of the penultimate layer of a neural network and to utilize the outputs up until the pen-ultimate layer (or any layer as a matter of fact) to estimate epistemic uncertainty. To accommodate the feature space for this, they state that it has to be both smooth and sensitive, because feature density alone might be a poor indicator for out-of-distribution (OoD) data and since density estimators might map the data to in-distribution (iD) regions in the feature space. To prevent this feature collapse, \cite{mukhoti2021deterministic} propose subjecting the feature extractor to a bi-Lipschitz constraint, such that 

\begin{align}
K_l d_I({x}, {x}') \le d_F(h({x}), h({x}')) \le K_u d_I({x}, {x}')
\end{align}

for all pairs of input data points ${x}, {x}' \in \mathcal{X}$, where $d_I$ and $d_F$ are distance measures for input space and feature space, respectively, $K_l$ and $K_u$ are Lipschitz constants, and $h$ is some smooth function. The constraint encourages sensitivity, i.e., preservation of distances in the input space through the lower bound and smoothness, i.e., prevention of too high sensitivity to input variations, through the upper bound. A simple approach to enforce a bi-Lipschitz constraint is via residual connections \citep{he2016deep} in a neural network model, e.g., as given in a conventional transformer architecture \citep{vaswani2017attention}, that is trained with spectral normalization \citep{miyato2018spectral} (see \cite{mukhoti2021deterministic} for details and references therein).

\subsection{Surjective normalizing flows}

Vanilla normalizing flows have previously been reported to fail to compute accurate likelihood estimates and detect OoD data \citep{kirichenko2020why, dai2020sliced,nalisnick2018deep,klein2021funnels}. To alleviate this shortcoming, we make use of a density estimator that projects the data into a lower dimensional manifold via surjective normalizing flows (SNF). We present the derivation thereof in the following section.

Following the SurVAE framework \citep{nielsen2020survae,klein2021funnels}, we model the log marginal likelihood $p({y})$ of a data point $y \in \mathbb{R}^p$ as

\begin{equation}
\log p({y}) = \log p({z}) + V({y}, {z}) + E({y}, {z}), \;\; {z} \sim q({z} | {y})
\label{eqn:survae}
\end{equation}

where $p(z)$ is a distribution that can be sampled from efficiently, $q({z} | {y})$ is some amortized distribution, $V({y}, {z})$ is denoted as \textit{likelihood contribution} and $E({y}, {z})$ is denoted as \textit{bound loeseness} term. Note that in this representation the random variable ${z}$ is latent while in the pushforward measure defining a typical normalizing flow, it is not. 

Similarly to \cite{klein2021funnels} and \cite{dirmeier2023simulation}, we partition the data $y$ into two components ${y} = [{y}^+, {y}^-]^T$ (with ${y}^+ \in \mathbb{R}^q$ and ${y}^- \in \mathbb{R}^{p - q}$) and define a \textit{generative} (dimensionality-increasing) transformation (i.e., from latent space to data space)

\begin{equation}
{y} = \begin{bmatrix}
{y}^{+} = f_{y^-}({z}) \\
{y}^{-} \sim p({y}^{-} | {z})
\end{bmatrix}
\label{eqn:generativetransformation}
\end{equation}

where $f_{y^-}: \mathbb{R}^q \rightarrow \mathbb{R}^q$ is a conditional normalizing flow \citep{winkler2019learning, papamakarios2021normalizing}, $f^{-1}_{y^-}(y^+)$ its inverse, and $p({y}^{-} | {z})$ is a conditional probability density function which we parameterize using a neural network. In the following, we drop the conditioning variable $y^-$ from $f_{y^-}$ for notational clarity.

To connect Equation~\eqref{eqn:survae} with the generative transformation of Equation~\eqref{eqn:generativetransformation}, we first observe that by composing a smooth function $g$ with a diffeomorphism $f$

\begin{align*}
    \int \delta \left( g(y)  \right) f\left( g(y) \right) \bigg| \det \frac{\partial g(y) }{\partial {y} }     \bigg| \mathrm{d}{y} = \int\delta (u) f(u) \mathrm{d}  {u} 
\end{align*}

one can conclude that

\begin{align*}
    \delta \left( g(y) \right)  = \delta (y - y_0) \bigg| \det \frac{\partial g \left( {y} \right) }{\partial {y} }     \bigg|^{-1}_{{y}={y}_0} 
\end{align*}

where $y_0$ is the root of $g(y)$ (see also \cite{nielsen2020survae,klein2021funnels,dirmeier2023simulation}). Making use of this derivation we define $g(y)= z - f^{-1}(y^+)$ (which has its root at $y_0 = f(z)$) and

\begin{align*}
q(z|y) &= \delta \left(z - f^{-1}(y^+) \right) = \delta(y - f(z)) |\det J(y^+)|^{-1}
\end{align*}

where $J(y^+) = \frac{\partial f^{-1}(y^+)}{\partial y^+}$ is the Jacobian of the bijective inverse transformation. We can now define the likelihood contribution as

\begin{align*}
\mathcal{V}({y}, {z}) &= \mathbb{E}_{q({z} | {y}) \rightarrow \delta({z} - f^{-1}({y}))} \left[ \log \frac{p({y} | {z})}{q({z} | {y})} \right] \\
&= \int \delta({z} - f^{-1}({y^+})) \log \frac{p(y^- | f^{-1}(y^+))}{ \delta({z} - f^{-1}({y^+})) |\det J(y^+)|^{-1}}   \mathrm{d}z \\
&= \int \delta(y^+ - f(z)) |\det J(y^+)|^{-1} \log \frac{p(y^- | z)}{ \delta(y^+ - f(z)) |\det J(y^+)|^{-1}  }   \mathrm{d}z \\
&= \int \delta(y^+ - \tilde{y}^+) \log \frac{p(y^- | z)}{ \delta(y^T - \tilde{y}^+) |\det J(y^+)|^{-1}  }  \mathrm{d} \tilde{y}^+ \\
&= \log p(y^- | f^{-1}(y^+))  + \log |\det J(y^+)|
\end{align*}

where we used the change of variables $\tilde{y}^+ = f(z)$ yielding $\mathrm{d}\tilde{y}^+ = \mathrm{d}z |\det J(y^+)|^{-1}$. Transferring this composition to Equation~\eqref{eqn:survae}:

\begin{equation}
\log p({y}) = \log  p \left( {z} \right) + \log p({y}^- | z) + \log \left| \det J(y^+) \right| + E({y}, {z})
\label{eqn:surjection}
\end{equation}

As \citet{nielsen2020survae} show, the bound looseness term $E({y}, {z})$ is zero in the case of inference surjections, i.e., the type of dimensionality reducing transformations which we are utilizing in Equation~\eqref{eqn:surjection}, or, more specifically, if the function $f^{-1}(y)$ satisfies the right inverse condition.

Normalizing flows are typically constructed by stacking several flow layers $f = \left(f_1, \dots,  f_K \right)$ which map a random variable from some easy-to-sample base distribution $z_0 \sim p(z_0)$ to some complex distribution $p(y)$ via the transformations $y = z_K = f_K \circ \dots \circ f_2 \circ f_1(z_0)$. The transformations increase flexibility and expressivity of the target distribution while retaining computational feasibility in case that the Jacobian determinants can be computed efficiently~\citep{rezende15nfvi,papamakarios2021normalizing}. 
%



For our model, we facilitate a mixture of dimensionality-retaining bijections and dimensionality-reducing surjections which results in a likelihood that consists of a) some base density term $p(z_0)$, b) Jacobian determinants $\det J_k(\cdot)$ contributed by conventional bijective normalizing flows, and c) Jacobian determinants $\det J_k(\cdot)$ plus conditional densities $p\left(z_k^-|f_k^{-1}(z_k^+) \right)$ contributed by surjective normalizing flows. Putting these together yields the following likelihood:

\begin{equation}
\begin{split}
\log p({y}) = & \ \log p({z}_0)
+ \sum_{k \in \mathcal{K}_b} \log \left| \det J_k(z_k) \right| \\
& + \sum_{k \in \mathcal{K}_s} \Big( \log p({z}^-_{k} \mid f^{-1}_k({z}^+_{k})) + \log \left| \det J_k(z_k) \right| \Big)
\end{split}
\end{equation}  

where $\mathcal{K}_b$ is a set of integers that index bijections and $\mathcal{K}_s$ indexes surjections, respectively and $\det J_k$ are the Jacobian determinants of the inverse mappings $f^{-1}_k$  (c.f. \cite{dirmeier2023simulation}).

Note that we generally distinguish two types of flows:
\begin{itemize}
    \item In the case of dimensionality-preserving bijections, the transformations $f^{-1}_k(z_k)$ are unconditional normalizing flows that act on the entire data vector $z_k$,
    \item In the case of dimensionality-reducing surjections, the transformations $f_k^{-1}(z_k^+) := f_{z_k^-}^{-1}(z_k^+)$ are conditional normalizing flows that act on a subset of $z_k$.
\end{itemize}

For OoD detection, we use a flow that combines bijective and surjective layers and apply it to estimate the density over the features $y \leftarrow h({x})$ derived from transforming the input data ${x}$ through a predictive neural network up to some feature layer. 

\section{Out-of-distribution detection using surjective normalizing flows for mobility research}
\label{sec:results}
The following section presents an application of the method in mobility research. We first present the data, then introduce the model and baselines, and finally demonstrate experimental results.

\subsection{Data}
\label{sec:data}

Following \cite{hong2022mobility}, we use simulated training data for validation of the method, since real data in mobility analysis usually underlies privacy protocols and cannot be shared for research. The training data consists of $N=800$ discrete location trajectories $\{ {y}_1, \dots, {y}_N \}$. Each trajectory ${y}_n = {y}_n^{1:2000} = (y_{n}^1, \dots, y_{n}^{2000})$ consists of $2000$ sequential location visits that have been generated through the density transition-EPR (DT-EPR) individual mobility model~\citep{hong2022mobility}. The model combines the density-EPR (d-EPR)~\citep{pappalardo_returners_2015} and individual preferential transition (IPT)~\citep{zhao_characteristics_2021} mechanisms, both derived from the exploration and preferential return (EPR) framework~\citep{song_modelling_2010}.
For $N$ simulated trajectories, the model begins by simulating an initial location $y_n^{(1)}$ uniformly from a set of available locations $\mathcal{L} = \{l_1, \dots, l_D\}$ where $D$ is the total number of selectable locations.
For each successive step, the agent that traverses the locations decides with probability $p_n^{(t + 1)} = \rho_n |{y}_n^{(1:t)}|^{- \gamma_n }$ to explore an unvisited location,  where $|{y}_n^{(1:t)}|$ denotes the number of unique locations in the trajectory up to ${y}_n^{(t)}$, and $\rho_n$ and $\gamma_n$ are fixed parameters denoting individual-specific exploration preferences (c.f. \cite{hong2022mobility} where they are sampled from Gaussians with parameters estimated from observational data).
Alternatively, the agent can choose to return to a previously visited location with probability $1 - p_n^{t + 1}$. 
In this case, EPR calculates the probability of selecting a known location as proportional to the visitation frequency of that location. 

In order to simulate OoD data and validate their models, \cite{hong2022mobility} compute interventional distributions by intervening on $\rho_n$, $\gamma_n$ and $p_n^{t}$ by two kinds of interventions:

\begin{itemize}
    \item They conduct shift interventions on the means of the distributions $\mathbb{P}(\rho)$ and $\mathbb{P}(\gamma)$, i.e., setting $\mathbb{P}(\rho) = \text{Normal}(\mu_{\rho}, \sigma^2_{\rho})$ with $\mu_{\rho} \in \{ 0.1, 0.4, 0.7, 0.9 \}$ and $\mathbb{P}(\gamma) = \text{Normal}(\mu_{\gamma}, \sigma^2_{\gamma})$ with $\mu_{\gamma} \in \{ 0.1, 0.4, 0.7, 0.9 \}$.
    \item They conduct hard interventions on the exploration probabilities $p_n^t$ by setting them to constants $p_n^t \in \{ 0.1, 0.25, 0.5, 0.75, 0.9\}$.
\end{itemize}

\cite{hong2022mobility} simulate "synthetic observational data", i.e., data where the distributions of random variables $\rho$ and $\gamma$ are parameterized via empirically estimated quantities $\hat{\mu}_{\gamma}$ and $\hat{\mu}_{\rho}$, and "synthetic interventional data", i.e., data where random variables are sampled from interventional distributions. They then fit a predictive model on a training set consisting of samples from the observational data set and evaluate the predictor on an observational test set and the interventional data sets.

\subsection{Model} 

We are interested in fitting a density estimator to be able to detect whether the typical trajectory of an individual shows motifs that have not been present in the training data and which would possibly render a prediction of a novel data point to be of low accuracy. 

We extend the transformer-based next location prediction model by \cite{hong_how_2022} to allow for quantification of epistemic and aleatoric uncertainty. 
In short, \cite{hong_how_2022} use the encoder architecture of a vanilla transformer model \citep{vaswani2017attention} to learn a predictive model for an individual's next location.
The predicted location is represented as a categorical variable, corresponding to specific places like supermarkets or workplaces, and is inferred from the previous location visit history of an individual.
Since the model by \cite{hong_how_2022} is already making use of residual connections, to guarantee a well-behaved feature space of the penultimate layer, one merely needs to spectral-normalize the weights of the attention matrices. 

We fit an SNF consisting of $10$ normalizing flow layers of which the third and eighth are dimensionality-reducing surjections. All flow layers (also the surjective ones) use masked coupling rational-quadratic neural spline flows (RQ-NSFs, \cite{dinh2014nice,dinh2016density,durkan2019neural,papamakarios2021normalizing}) as inverse transformations, i.e., from ambient space to embedding space, using $64$ bins defined on an interval from $-5.0$ to $5.0$ to fit the boundaries of the data. We use MLP conditioners with $2$ hidden layers and $512$ nodes per layer to compute the parameters of the RQ-NSFs. The surjective layers require an additional decoder, to parameterize the conditional distribution $p_\phi \left( {y}^{-} \mid f^{-1}_{{y}^-} \left( {y}^+\right) \right)$ for which we also use an MLP with $2$ hidden layers with $512$ nodes each. We use conditional multivariate Gaussians for $p_\phi$ for which the mean and diagonal of the covariance matrix are computed with the former MLP.

As baselines we use a fully bijective normalizing flow (BNF) that is parameterized as the model described above but does not make use of surjective layers. As a second baseline, we fit a Dirichlet process Gaussian mixture model (DPGMM). We truncate the DPGMM at a maximum of $50$ mixture components, such that we can use an optimization-based variational approach for Bayesian inference (see Appendix~\ref{appendix:experimental-details} for experimental details).

\subsection{Experimental results}

We fit the three models to the observational training data set of Section~\ref{sec:data} and evaluate their likelihood estimates on both observational training and test data sets for model selection (Table~\ref{table:likelihoods}). Despite its complexity, the DPGMM achieves significantly worse likelihood estimates on both train and test sets while the bijective flow (BNF) and the surjective flow (SNF) are relatively similar. We then evaluated the three models on the interventional data sets and computed a likelihood estimate for each data point and model. 

\begin{table*}[h!]
\centering
\begin{tabular}{@{}lrr@{}}
\toprule
& Train set & Test set \\
\midrule
DPGMM  & $147.5$9 & $147.52$ \\
Bijective normalizing flow  & $43.33$ & $49.57$ \\
Surjective normalizing flow & $\mathbf{41.46}$ & $\mathbf{47.25}$ \\
\bottomrule
\end{tabular}
\caption{Negative log-likelihoods on training set (including validation set) and test set for all three models (lower is better). The flow model that uses surjections achieves more competitive estimates than the other two models.}
\label{table:likelihoods}
\end{table*}

\begin{figure}
\includegraphics[width=1\textwidth]{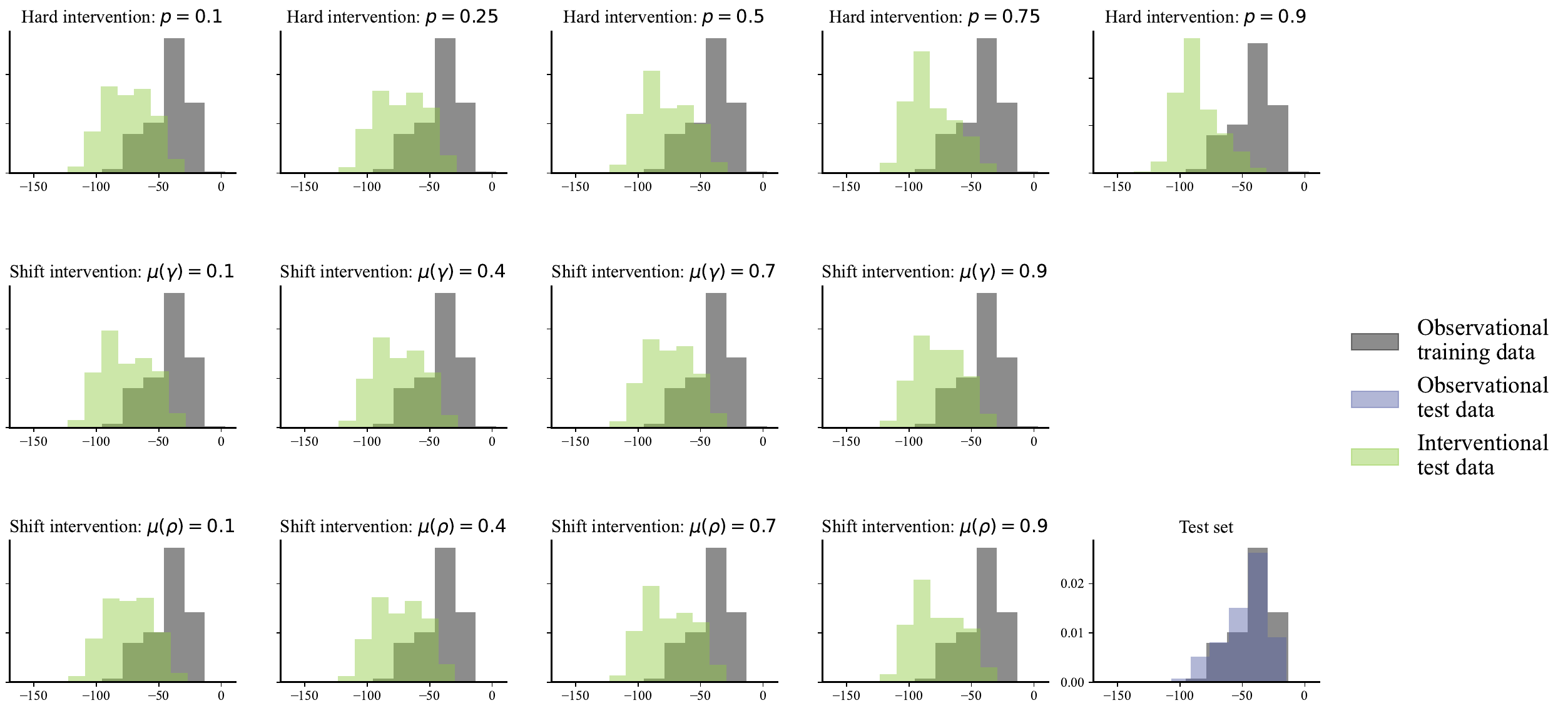}
\caption{Out-of-distribution detection via density estimation. The SNF manages to reliably detect out-of-distribution data either using simple graphical checks or by conducting statistical tests between pairs of distributions of likelihood estimates. Each of the OoD data sets (green) has a significant distributional shift from the training data set(grey). On the test data set (blue) no significant difference is visible.}
\label{figfigure:oodd}
\end{figure}

Graphically, the SNF manages to more reliably detect if a distribution has a distributional shift w.r.t. the training distribution than the other two methods (Figure~\ref{figfigure:oodd} for SNF and Appendix~\ref{appendix:additional-results} for the same visualizations for BNF and DPGMM). In contrast, note the strong overlap between the test data and training data likelihood distributions.

\begin{table*}
\centering
\begin{tabular}{@{}l rrr r rrr@{}}
\toprule
 & \multicolumn{3}{c}{$N=100$} && \multicolumn{3}{c}{$N=200$}\\
\cmidrule{2-4}\cmidrule{6-8}
&     DPGMM &     BNF &    SNF &&     DPGMM &     BNF &    SNF \\
\midrule
Train set                                &   $1.00$ &  $1.00$ &  $1.00$ &&  $1.00$ &  $1.00$ &  $1.00$ \\
Test set                                 &  $\mathbf{0.47}$ &  $0.07$ &   ${0.09}$ &&  $\mathbf{0.56}$ &  $8.05^{-3}$ &  ${1.20^{-2}}$ \\
\cmidrule{1-8}
Intervention: $p=0.5$               &  $0.47$ &  $3.56^{-11}$ &  $\mathbf{9.23^{-25}}$ &&  $0.39$ &  $4.52^{-22}$ &  $\mathbf{1.43^{-47}}$ \\
Intervention: $\mu(\gamma)= 0.4$   &  $0.51$ &  $2.80^{-9}$ &   $\mathbf{2.15^{-21}}$ &&  $0.47$ &  $6.40^{-21}$ &  $\mathbf{1.36^{-45}}$ \\
Intervention: $\mu(\rho) = 0.4$    &  $0.49$ &  $9.41^{-9}$ &  $\mathbf{4.60^{-20}}$ &&  $0.49$ &  $2.92^{-19}$ &  $\mathbf{1.46^{-44}}$ \\
\bottomrule
\end{tabular}
\caption{Two-sample $t$-tests. Table shows average $p$-values of 100 repetitions of testing for a difference in means between two randomly sub-sampled feature space likelihood estimates of different data sets. For the first two rows high $p$-values are better, for the last three rows low $p$-values are better (see Appendix~\ref{appendix:additional-results} for full table and results with Wasserstein distance.)}
\label{table:pvalueresults}
\end{table*}

More quantitatively, one can detect if a data set is OoD by, for instance, testing if the means of sub-samples of the feature space likelihood estimates of two data set are significantly separated from each other (see an excerpt in Table~\ref{table:pvalueresults} and Appendix~\ref{appendix:additional-results} for additional results).

\section{Conclusion}
\label{sec:conclusion}
We introduced a simple method for out-of-distribution detection for data sets from mobility analysis by computing epistemic uncertainty estimates in a deep neural network. The only requirement the method has is to appropriately normalize weights in a neural network architecture during training and to add residual connections to the network structure. We detect OoD samples by comparing the distribution of their epistemic uncertainty estimates to the distribution of uncertainty estimates of the training data and "reject" a sample to be OoD based on a simple statistical significance test. If the test fails to reject the null, we conclude the sample to be iD. We showed that this simple approach can convincingly detect OoD samples on a synthetic observational data set and a multitude of interventional data sets. 

Uncertainty quantification (UQ) is of central importance in a wide variety of applications ranging from research in biology to finance to mobility research. Future research in mobility could focus on how ideas from causality, for instance, distributional robustness via invariant prediction or conditional variance penalities \citep{peters2016causal,heinze2021conditional} can be integrated into DNN models that already are capable of UQ. Reliable UQ and distributional robustness could pose promising future research avenues to further establish deep learning in application-heavy fields where accurate decision making is crucial.

\section*{Acknowledgments and Disclosure of Funding}

This work was supported by the Hasler Foundation under the project titled "Interpretable and Robust Machine Learning for Mobility Analysis" (grant number 21041).

\newpage
\bibliographystyle{plainnat}
\bibliography{references} 

\newpage
\begin{appendices}
\section{Experimental details}
\label{appendix:experimental-details}

To produce the experimental results in Section~\ref{sec:results}, we trained a surjective normalizing flow (SNF), a bijective normalizing flow (BNF) as well as a Dirichlet process Gaussian mixture model (DPGMM). We delineate the architectural choices, training procedures, etc. in detail below.

\subsection{Normalizing flow models}

The SNF uses $10$ normalizing flow layers that consist of dimensionality-reducing surjections at the $3$rd and $8$th layer and dimensionality-preserving bijections everywhere else. The two surjective layers reduce the dimensionality by $25\%$ each meaning that a random subset consisting of $25\%$ of the input vector was randomly chosen to be discarded. In order to not discard information of a relevant dimension it sufficies to use bijective layers in front and intermingle them with surjective layers. The architecture of the SNF was chosen somewhat arbirarily and no hyperparameterization optimization was conducted for a more objective assessment of the model. 

Irrespective of the type of layer, all layers use mask coupling with RQ-NSFs \citep{dinh2014nice,dinh2016density,durkan2019neural} as forward transformations $f$ using two-layer MLPs with $512$ nodes as conditioners (see \cite{papamakarios2021normalizing} for denotations). RQ-NSFs use half of the input vector as conditioning variables and output parameters for a RQ-spline with $64$ bins defined on the interval $[-5, 5]$. The conditional distributions $p(\bm{y}^-|\bm{z})$ are parameterized as conditional Gaussians using MLPs consisting of two layers with $512$ hidden nodes each. All neural networks used \texttt{relu} activation functions. 

We chose the BNF to be identical in architecture as the SNF with the exception that all layers are dimensionality-preserving. Specifically, all layers are masked coupling layers that use RQ-NSFs and MLP conditioners using two layers of $512$ nodes.

We train both models until convergence on mini-batches of size $128$ using an AdamW optimizer \citep{loshchilov2018decoupled} for training  with a learning rate of $l=0.0003$ and otherwise default parameters using the Python library \texttt{Optax} \citep{deepmind2020jax}.

Both SNF and BNF are implemented using Python library \texttt{Surjectors} \citep{dirmeier2023surjectors}.

\subsection{Mixture model}

We fit DPGMMs with truncation at $K \in \{25, 50, 75, 100\}$ components using the following generative model

\begin{equation*}
\begin{split}
    \alpha & \sim \text{Gamma}(1, 1)\\
    \nu_k &\sim \text{Beta}(1, \alpha) \\
    \boldsymbol \pi &\leftarrow \text{break-stick}(\boldsymbol \nu)\\
    \boldsymbol \mu_{k} & \sim \text{MvNormal}(\bm{0}, \bm{1})\\
    \sigma_{kp} & \sim \text{HalfNormal}(1)\\
    \Sigma_k & \leftarrow \text{diag}(\boldsymbol \sigma^2_k) \\    
    y_i & \sim \sum_{k=1}^K \pi_k \cdot \text{MvNormal}(\boldsymbol \mu_k, \boldsymbol \Sigma_k)
\end{split}
\end{equation*}

where we characterize the DP using the (truncated) stick-breaking construction \citep{ishwaran2001gibbs,blei2006var}.

\begin{equation*}
\pi_i(\boldsymbol \nu) = \nu_i \prod_{j=1}^{i - 1} (1 - \nu_j)
\end{equation*}

which is amendable to optimization using automatic-differentiation variational inference \citep{hoffman2013stochastic,kucukelbir2017automatic} which we conduct using the probabilistic programming language \texttt{NumPyro} \citep{phan2019composable} using an Adam optimizer with learning rate $lr=0.003$ \citep{kingma2015adam}.

We compute the likelihood of the test data $\bm{y}^*$ via integrating over the posterior of the variables $\boldsymbol \theta = (\pi_k, \boldsymbol \mu_k, \boldsymbol \Sigma_k)$: $\int p(\bm{y}^* | \boldsymbol \theta)  p(\boldsymbol \theta | \bm{y}) d\boldsymbol \theta$.

\section{Additional experimental results}
\label{appendix:additional-results}

\subsection{Source code}

Source code to fit a custom density estimator for OoD detection can be found on \href{https://github.com/irmlma/uncertainty-quantification-snf}{GitHub}.

\subsection{Additional figures}

\begin{figure}[h!]
\begin{subfigure}[b]{1\textwidth}
\includegraphics[width=1\textwidth]{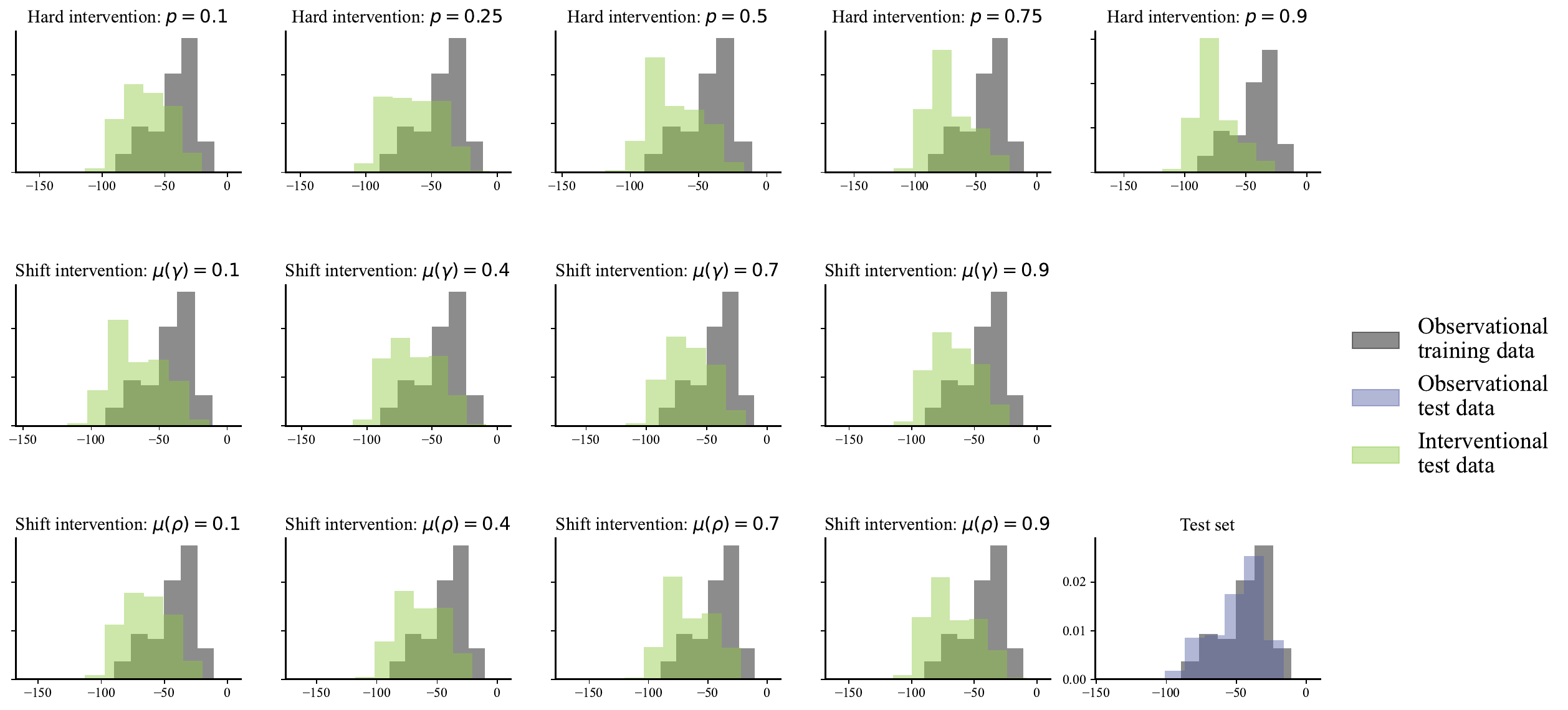}
\caption{Out-of-distribution detection BNF. We visualize the empirical density estimates when using a BNF as a density estimator (compare Figure~1).}

\end{subfigure}
\begin{subfigure}[b]{1\textwidth}
\includegraphics[width=1\textwidth]{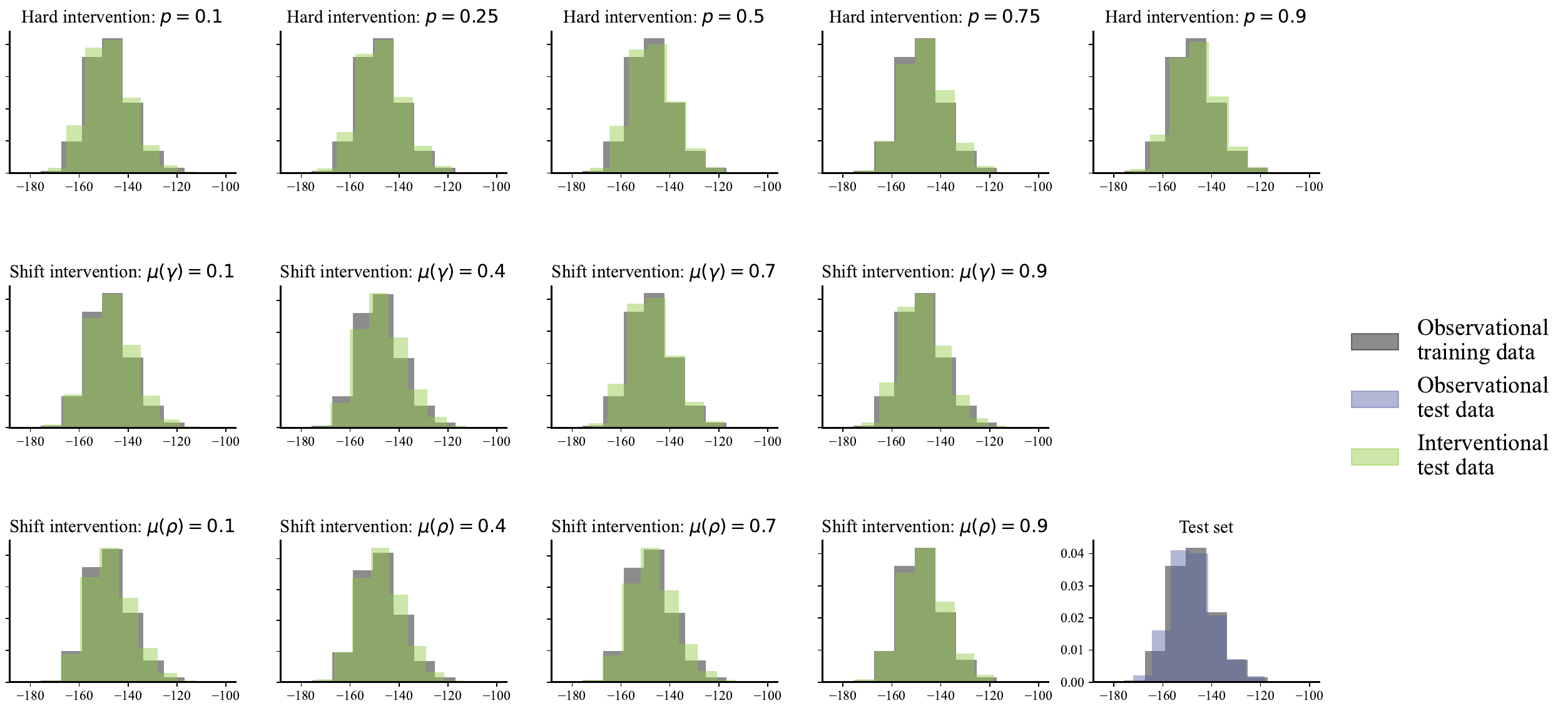}
   \caption{Out-of-distribution detection DPGMM. We visualize the empirical density estimates when using a DPGMM as a density estimator (compare Figure~1).}
\end{subfigure}
\label{fig:ood-detection}
\end{figure}

\begin{table*}
\centering
\begin{tabular}{@{}l rrr r rrr@{}}
\toprule
 & \multicolumn{3}{c}{$N=100$} && \multicolumn{3}{c}{$N=200$}\\
\cmidrule{2-4}\cmidrule{6-8}
&     DPGMM &     BNF &    SNF &&     DPGMM &     BNF &    SNF \\
\midrule
Train set                                & $1.00$ &  $1.00$ & $1.00$ &&    $1.00$ &  $1.00$ &  $1.00$ \\
Test set                                 &  $0.47$ &    $0.07$ &    $0.09$ &&  $0.56$ &   $8.05^{-03}$ &  $1.20^{-02}$ \\
\cmidrule{1-8}
Hard intervention: $p=0.1$               & $0.50$ & $2.73^{-8}$ & $1.84^{-19}$ && $0.49$ &  $6.40^{-20}$ & $1.77^{-46}$ \\
Hard intervention: $p=0.25$              &   $0.51$ & $2.90^{-8}$ &  $7.03^{-19}$ &&  $0.53$ &  $8.65^{-15}$ & $1.42^{-37}$ \\
Hard intervention: $p=0.5$               & $0.47$ & $3.56^{-11}$ &  $9.23^{-25}$ &&  $0.39$ &  $4.52^{-22}$ & $1.43^{-47}$ \\
Hard intervention: $p=0.75$              & $0.44$ &   $1.61^{-16}$ &  $2.00^{-30}$ &&  $0.37$ &  $3.22^{-36}$ &  $1.48^{-63}$ \\
Hard intervention: $p=0.9$               & $0.35$ &  $3.95^{-21}$ &   $5.66^{-37}$ && $0.294$ &  $2.03^{-45}$ &  $8.66^{-78}$ \\
Shift intervention: $\mu ( \gamma) = 0.1$ &  $0.47$ &   $4.00^{-11}$ &  $3.82^{-24}$ &&  $0.46$ & $2.34^{-21}$ &  $9.17^{-48}$ \\
Shift intervention: $\mu ( \gamma) = 0.4$ &   $0.51$ &  $2.80^{-9}$ & $2.15^{-21}$ &&  $0.47$ & $6.40^{-21}$ & $1.36^{-45}$ \\
Shift intervention: $\mu ( \gamma) = 0.7$ &  $0.53$ &  $9.00^{-10}$ &  $1.44^{-23}$ && $0.49$ &  $5.15^{-21}$ &  $7.80^{-48}$ \\
Shift intervention: $\mu ( \gamma) = 0.9$ &  $0.50$ &   $4.70^{-12}$ & $1.49^{-25}$ && $0.44$ & $1.14^{-20}$ &  $5.41^{-46}$ \\
Shift intervention: $\mu ( \rho) = 0.1$   &  $0.54$ &  $8.43^{-9}$ &  $8.30^{-22}$ &&  $0.53$ & $2.11^{-19}$ &  $3.91^{-46}$ \\
Shift intervention: $\mu ( \rho) = 0.4$   & $0.49$ &  $9.41^{-9}$ &  $4.60^{-20}$ &&  $0.49$ & $2.92^{-19}$ & $1.46^{-44}$ \\
Shift intervention: $\mu ( \rho) = 0.7$   & $0.49$ &   $4.23^{-10}$ & $4.22^{-22}$ &&  $0.48$ & $1.04^{-21}$ &  $7.44^{-46}$ \\
Shift intervention: $\mu ( \rho) = 0.9$   & $0.48$ &  $1.32^{-10}$ & $1.38^{-21}$ && $0.42$ & $1.14^{-25}$ & $1.27^{-49}$ \\

\bottomrule
\end{tabular}
\caption{Two-sample $t$-tests. Table shows average $p$-values of 100 repetitions of testing for a difference in means between two randomly sub-sampled feature space likelihood estimates of different data sets. For the first two rows high $p$-values are better, for the last rows low $p$-values are better.}
\label{table:all-wasserstein-table}
\end{table*}

\begin{table*}
\centering
\begin{tabular}{@{}l rrr r rrr@{}}
\toprule
 & \multicolumn{3}{c}{$N=100$} && \multicolumn{3}{c}{$N=200$}\\
\cmidrule{2-4}\cmidrule{6-8}
&     DPGMM &     BNF &    SNF &&     DPGMM &     BNF &    SNF \\
\midrule
Train set                                &        $0.00$ &      $0.00$ &       $0.00$ &&  $0.00$ &      $0.00$ &       $0.00$ \\
Test set                                 &        $1.62$ &      $6.58$ &       $6.04$ &&  $1.07$ &      $6.56$ &       $6.03$ \\
\cmidrule{1-8}
Hard intervention: $p=0.1$               &        $1.53$ &     $20.88$ &      $32.79$ &&  $1.09$ &     $21.10$ &      $33.08$ \\
Hard intervention: $p=0.25$              &        $1.63$ &     $19.80$ &      $31.31$ &&  $1.10$ &     $20.05$ &      $31.80$ \\
Hard intervention: $p=0.5$               &        $1.69$ &     $23.94$ &      $35.70$ &&  $1.29$ &     $24.08$ &      $35.83$ \\
Hard intervention: $p=0.75$              &        $1.82$ &     $28.54$ &      $40.60$ &&  $1.36$ &     $28.46$ &      $40.32$ \\
Hard intervention: $p=0.9$               &        $2.04$ &     $32.35$ &      $44.44$ &&  $1.59$ &     $32.35$ &      $44.54$ \\
Shift intervention: $\mu ( \gamma) = 0.1$ &        $1.75$ &     $22.45$ &      $34.35$ &&  $1.21$ &     $22.72$ &      $34.59$ \\
Shift intervention: $\mu ( \gamma) = 0.4$ &        $1.64$ &     $21.08$ &      $33.13$ &&  $1.18$ &     $21.32$ &      $33.26$ \\
Shift intervention: $\mu ( \gamma) = 0.7$ &        $1.53$ &     $21.59$ &      $33.38$ &&  $1.15$ &     $21.53$ &      $33.47$ \\
Shift intervention: $\mu ( \gamma) = 0.9$ &        $1.59$ &     $22.62$ &      $34.49$ &&  $1.23$ &     $22.65$ &      $34.48$ \\
Shift intervention: $\mu ( \rho) = 0.1$   &        $1.61$ &     $20.40$ &      $32.39$ &&  $1.10$ &     $20.89$ &      $32.71$ \\
Shift intervention: $\mu ( \rho) = 0.4$   &        $1.61$ &     $20.54$ &      $32.32$ &&  $1.12$ &     $20.62$ &      $32.25$ \\
Shift intervention: $\mu ( \rho) = 0.7$   &        $1.63$ &     $22.48$ &      $34.33$ &&  $1.17$ &     $22.37$ &      $34.16$ \\
Shift intervention: $\mu ( \rho) = 0.9$   &        $1.72$ &     $23.81$ &      $35.60$ &&  $1.28$ &     $24.20$ &      $36.13$ \\
\bottomrule
\end{tabular}
\caption{Two-sample $1d$-Wasserstein distances. Table shows average Wasserstein distances of 100 repetitions between two randomly sub-sampled feature space likelihood estimates of different data sets. For the first two rows high low distances are better, for the last rows high distances are better.}
\label{table:all-wasserstein-table}
\end{table*}

\end{appendices}

\end{document}